\pdfoutput=1

\documentclass[11pt]{article}

\usepackage{emnlp2021}

\usepackage{times}
\usepackage{latexsym}

\usepackage[T1]{fontenc}

\usepackage[utf8]{inputenc}

\usepackage{microtype}

\usepackage{url}
\usepackage{graphicx}
\usepackage{subcaption}
\usepackage{amsfonts}
\usepackage{multirow}
\usepackage{makecell}

\usepackage{mathrsfs}

\usepackage{amsmath,amssymb}

\usepackage{algorithm}
\usepackage{algorithmic}
\usepackage[english]{babel}
\usepackage[utf8]{inputenc}

\title{WeChat Neural Machine Translation Systems for WMT21}

\author{\text{Xianfeng Zeng}\textsuperscript{1}\thanks{\quad Equal contribution.} , \text{Yijin Liu}\textsuperscript{12}\footnotemark[1] , \text{Ernan Li}\textsuperscript{1}\footnotemark[1] , \text{Qiu Ran}\textsuperscript{1}\footnotemark[1] , \text{Fandong Meng}\textsuperscript{1}\footnotemark[1] , \\ 
\textbf{Peng Li}\textsuperscript{1}, \textbf{Jinan Xu}\textsuperscript{2}, \textbf{and Jie Zhou}\textsuperscript{1} \\
\textsuperscript{1} \text{Pattern Recognition Center, WeChat AI, Tencent Inc, China} \\
\textsuperscript{2} \text{Beijing Jiaotong University, Beijing, China} \\
{\{xianfzeng,yijinliu,cardli,soulcaptran,fandongmeng,patrickpli,withtomzhou\}@tencent.com} \\
{jaxu@bjtu.edu.cn}
}

\date{}

\begin{document}
\maketitle
\begin{abstract}
This paper introduces WeChat AI's participation in WMT 2021 shared news translation task on  English$\to$Chinese, English$\to$Japanese, Japanese$\to$English and English$\to$German. Our systems are based on the Transformer \cite{transformer2017} with several novel and effective variants. In our experiments, we employ data filtering, large-scale synthetic data generation (i.e., back-translation, knowledge distillation, forward-translation, iterative in-domain knowledge transfer), advanced finetuning approaches, and boosted Self-BLEU based model ensemble. Our \textbf{constrained} systems achieve 36.9, 46.9, 27.8 and 31.3 case-sensitive BLEU scores on English$\to$Chinese, English$\to$Japanese, Japanese$\to$English and English$\to$German, respectively. The BLEU scores of English$\to$Chinese, English$\to$Japanese and Japanese$\to$English are the highest among all submissions, and that of English$\to$German is the highest among all constrained submissions. 
\end{abstract}

\section{Introduction} \label{introduction}
We participate in the WMT 2021 shared news translation task in three language pairs and four language directions, English$\to$Chinese, English$\leftrightarrow$Japanese, and English$\to$German. In this year's translation tasks, we mainly improve the final ensemble model's performance by increasing the diversity of both the model architecture and the synthetic data, as well as optimizing the ensemble searching algorithm.

Diversity is a metric we are particularly interested in this year. To quantify the diversity among different models, we compute Self-BLEU \cite{selfbleu2018texygen} from the translations of the models on the valid set. To be precise, we use the translation of one model as the hypothesis and the translations of other models as references to calculate an average BLEU score. A higher Self-BLEU means this model is less diverse. 

For model architectures~\cite{transformer2017,meng2019dtmt,yanetal2020multi}, we exploit several novel Transformer variants to strengthen model performance and diversity. 
Besides the Pre-Norm Transformer, the Post-Norm Transformer is also used as one of our baselines this year. We adopt some novel initialization methods \cite{huang2020improving} to alleviate the gradient vanishing problem of the Post-Norm Transformer.
We combine the Average Attention Transformer (AAN) \cite{zhangetalaan2018} and Multi-Head-Attention \cite{transformer2017} to derive a series of effective and diverse model variants.
Furthermore, Talking-Heads Attention \cite{shazeer2020talking} is introduced to the Transformer and shows a significant diversity from all the other variants.

For the synthetic data generation, we exploit the large-scale back-translation \cite{sennrichetalimproving2016} method to leverage the target-side monolingual data and the sequence-level knowledge distillation \cite{kim-rush-2016-sequence} to leverage the source-side of bilingual data. 
To use the source-side monolingual data, we explore forward-translation by ensemble models to get general domain synthetic data. We also use iterative in-domain knowledge transfer \cite{meng2020wechat} to generate in-domain data. 
Furthermore, several data augmentation methods are applied to improve the model robustness, including different token-level noise and dynamic top-p sampling.

For training strategies, we mainly focus on scheduled sampling based on decoding steps \cite{liuemnlp2021scheduled}, the confidence-aware scheduled sampling \cite{mihaylovamartinsscheduled2019,parallel2019schedule,liu2021confidence}, the target denoising \cite{meng2020wechat} method and the Graduated Label Smoothing \cite{wang2020inference} for in-domain finetuning. 

For model ensemble, we select high-potential candidate models based on two indicators, namely model performance (BLEU scores on valid set) and model diversity (Self-BLEU scores among all other models). Furthermore, we propose a search algorithm based on the Self-BLEU scores between the candidate models with selected models.
We observed that this novel method can achieve the same BLEU score as the brute force search while saving approximately 95\% of search time.

This paper is structured as follows: Sec.~\ref{architectures} describes our novel model architectures. We present the details of our systems and training strategies in Sec.~\ref{techniques}. Experimental settings and results are shown in Sec.~\ref{experiments}. We conduct analytical experiments in Sec.~\ref{analysis}. Finally, we conclude our work in Sec.~\ref{conclusion}.

\section{Model Architectures} \label{architectures}

In this section, we describe the model architectures used in the four translation directions, 
including several different variants for the Transformer \cite{transformer2017} .

\subsection{Model Configurations}
Deeper and wider architectures are used this year since they show strong capacity as the number of parameters increases. In our experiments, we use multiple model configurations with 20/25-layer encoders for deeper models and the hidden size is set to 1024 for all models. Compared to our WMT20 models \cite{meng2020wechat}, we also increase the decoder depth from 6 to 8 and 10 as we find that gives a certain improvement, but deeper depths give limited performance gains. For the wider models, we adopt 8/12/15 encoder layers and 1024/2048 for hidden size. The filter sizes of models are set from 8192 to 15000. Note that all the above model configurations are applied to the following variant models.

\subsection{Transformer with Different Layer-Norm}
The Transformer \cite{transformer2017} with Pre-Norm \cite{xiong2020on} is a widely used architecture in machine translation. It is also our baseline model as its performance and training stability is better than the Post-Norm counterpart.

Recent studies \cite{liu2020very,huang2020improving} show that the unstable training problem of Post-Norm Transformer can be mitigated by modifying initialization of the network and the successfully converged Post-Norm models generally outperform Pre-Norm counterparts. 
We adopt these initialization methods \cite{huang2020improving} to our training flows to stabilize the training of deep Post-Norm Transformer. Our experiments have shown that the Post-Norm model has a good diversity compared to the Pre-Norm Model and slightly outperform the Pre-Norm Model. We will further analyze the model diversity of different variants in Sec.~\ref{effect_of_model}.

\subsection{Average Attention Transformer}
We also use Average Attention Transformer (AAN) \cite{zhangetalaan2018} as we used last year to introduce more model diversity. In the Average Attention Transformer, a fast and straightforward average attention is utilized to replace the self-attention module in the decoder with almost no performance loss. The context representation $g_i$ for each input embedding is as follows:
\begin{equation}
    g_i = FFN(\frac{1}{i} \sum_{k=1}^i y_k)
\end{equation}
where $y_k$ is the input embedding for step $k$ and $i$ is the current time step. $FFN(\cdot)$ denotes the position-wise feed-forward network proposed by \citet{transformer2017}.

In our preliminary experiments, we observe that the Self-BLEU \cite{selfbleu2018texygen} scores between AAN and Transformer are lower than the scores between the Transformer with different configurations.

\subsection{Weighted Attention Transformer}
We further explore three weighting strategies to improve the modeling of history information from previous positions in AAN. Compared to the average weight across all positions, we try three methods including decreasing weights with position increasing, learnable weights and exponential weights. In our experiments, We observe exponential weights perform best among all these strategies. The exponential weights context representation $g_i$ is calculated as follows:
\begin{equation}
\setlength{\abovedisplayskip}{3pt}
\setlength{\belowdisplayskip}{3pt}
   c_i = (1-\alpha)y_i + \alpha \cdot c_{i-1}
\end{equation}
\begin{equation}
\setlength{\abovedisplayskip}{3pt}
\setlength{\belowdisplayskip}{3pt}
    g_i = FFN(c_i)
\end{equation}
where $\alpha$ is a tuned parameter. In our previous experiments, we test different alpha, including 0.3, 0.5, and 0.7, on the valid set and we set the alpha to 0.7 in all subsequent experiments as it slightly outperform the others.
\begin{figure}[t!]
\begin{center}
      \includegraphics[width=0.45\textwidth]{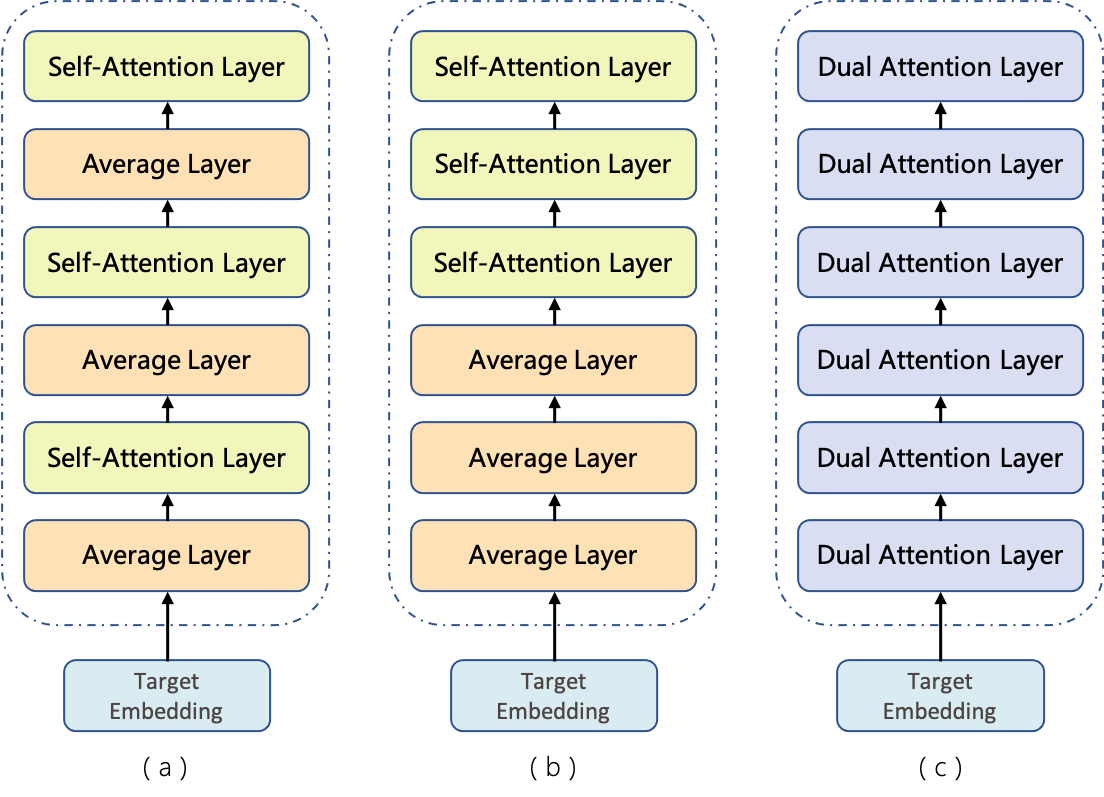}
      \caption{Mixed-AAN Transformers. } \label{f:mixaan}
\end{center}
\end{figure}
\subsection{Mixed-AAN Transformers}
Our preliminary experiments show that the decoder structure is strongly related to the model diversity in the Transformer. Therefore, we propose to stack different types of decoder layers to derive different Transformer variants. As shown in Figure \ref{f:mixaan}, we mainly adopt three Mixed-AAN Transformer architectures: a) Alternately mixing the standard self-attention layer and the average attention layer, b) Continuously stacking several average attention layers on the bottom layers and then stacking self-attention layers for the rest layers. c) Stacking both the self-attention layer and average attention layer at each layer and using their average sum to form the final hidden states (named as `dual attention layer').

In the experiments, Mixed-AAN not only performs better but also shows strong diversity compared to the vanilla Transformer. With four Mixed-AAN models, we reach a better ensemble result than the result with ten models which consist of deeper and wider standard Transformer. We will further analyze the effects of different architectures from performance, diversity, and model ensemble in Sec.~\ref{effect_of_model}

\subsection{Talking-Heads Attention}
In Multi-Head Attention, the different attention heads perform separate computations,  which are then summed at the end. Talking-Heads Attention \cite{shazeer2020talking} is a new variation that inserts two additional learned linear projection weights, $W_l$ and $W_w$, to transform the attention-logits and the attention scores respectively, moving information across attention heads. 
The calculation formula is as follows:
\begin{equation}
\hspace{-0.65 mm}
 \resizebox{.9\hsize}{!}{\, $Attention(Q, K, V) = softmax (\frac{QK^T}{\sqrt{d_k}} W_l) W_w V$}
\end{equation}

We adopt this method in both encoders and decoders to improve information interaction between attention heads. This approach shows the most remarkable diversity among all the above variants with only a slight performance loss.

\begin{figure*}[t!]
\begin{center}
      \includegraphics[width=0.9\textwidth]{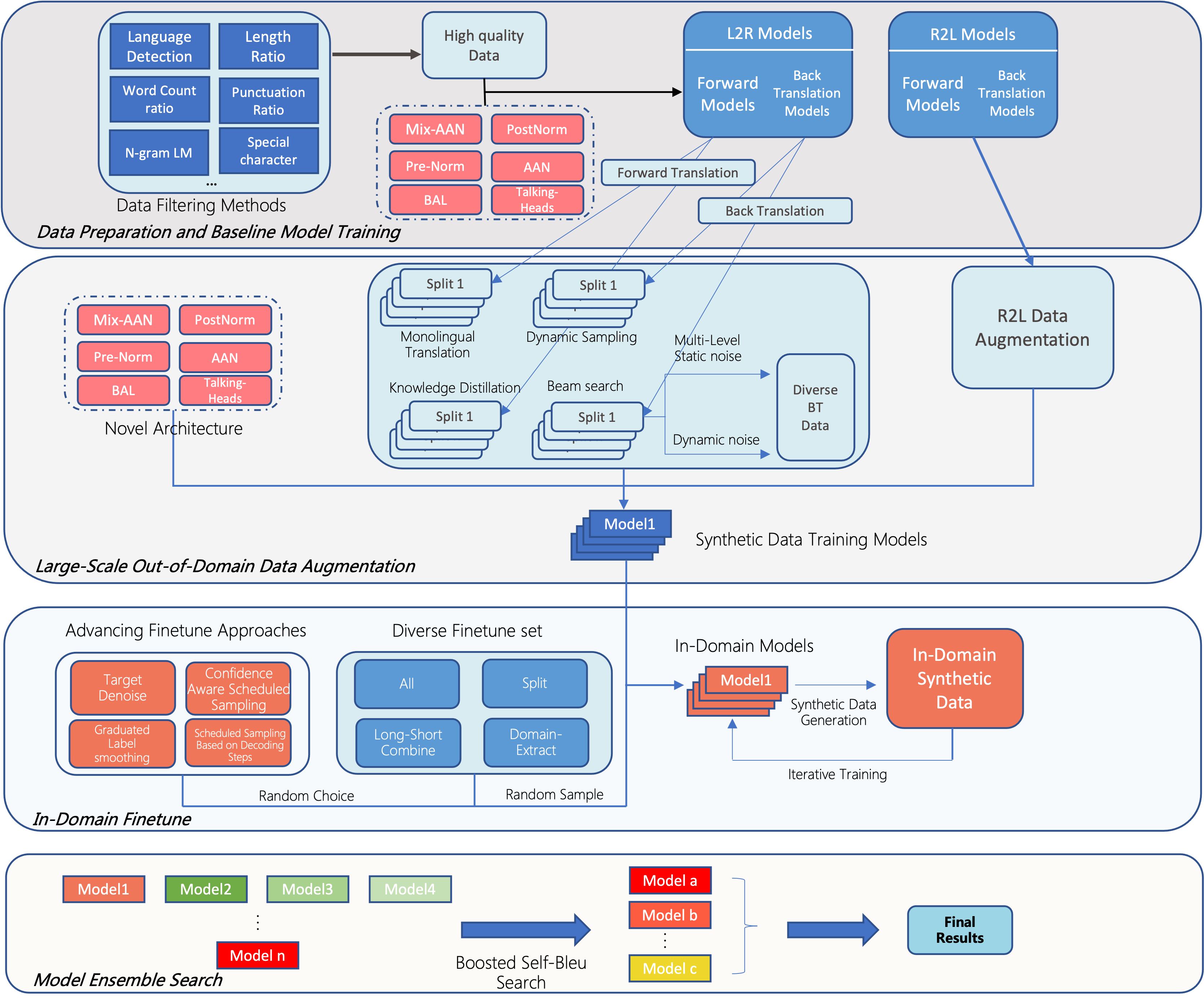}
      \caption{Architecture of our NMT system. } \label{f:main_graph}
 \end{center}
\end{figure*}

\section{System Overview}  \label{techniques}
In this section, we describe our system used in the WMT 2021 news shared task. 
We depicts the overview of our NMT system in Figure \ref{f:main_graph}, which can be divided into four parts, namely data filtering, large-scale synthetic data generation, in-domain finetuning, and ensemble. The synthetic data generation part further includes the generation of general domain and in-domain data. Next, we proceed to illustrate these four parts.

\subsection{Data Filtering}
\label{sec:data_filter}
We filter the bilingual training corpus with the following rules for most language pairs:
\begin{itemize}
\item Normalize punctuation with Moses scripts except Japanese data.
\item Filter out the sentences longer than 100 words or exceed 40 characters in a single word.
\item Filter out the duplicated sentence pairs.
\item The word ratio between the source and the target words must not exceed 1:4 or 4:1.
\item Filter out the sentences where the fast-text result does not match the origin language.
\item Filter out the sentences that have invalid Unicode characters.
\end{itemize}
Besides these rules, we filter out sentence pairs in which Chinese sentence has English characters in En-Zh parallel data.
The monolingual corpus is also filtered with the n-gram language model trained by the bilingual training data for each language. All the above rules are applied to synthetic parallel data.

\subsection{General Domain Synthetic Data Generation}
In this section, we describe our techniques for constructing general domain synthetic data. 
The general domain synthetic data is generated via large-scale back-translation, forward-translation and knowledge distillation to enhance the models' performance for all domains. 
Then, we exploit the iterative in-domain knowledge transfer \cite{meng2020wechat} in Sec \ref{sec:ind-trans}, which transfers in-domain knowledge to the vast source-side monolingual corpus, and builds our in-domain synthetic data.
In the following sections, we elaborate the above techniques in detail.

\subsubsection{Large-scale Back-Translation}

Back-translation is the most commonly used data augmentation technique to incorporate the target side monolingual data into NMT \cite{hoang2018iterative}. Previous work \cite{edunov-etal-2018-understanding} has shown that different methods of generating pseudo corpus has a different influence on translation quality. Following these works, we attempt several generating strategies as follows:
\begin{itemize}
\item Beam Search: Generate target translation by beam search with beam size 5.
\item Sampling Top-K: Select a word randomly from top-K (K is set to 10) words at each decoding step.
\item Dynamic Sampling Top-p: Selected a word at each decoding step from the smallest set whose cumulative probability mass exceeds p and the p is dynamically changing from 0.9 to 0.95 during data generation.
\end{itemize}

Note that we also use Tagged Back-Translation \cite{caswell-etal-2019-tagged} in En$\to$De and Right-to-Left (R2L) back-translation in En$\leftrightarrow$Ja, as we achieve a better BLEU score after using these methods.

\subsubsection{Knowledge Distillation}

Knowledge Distillation (KD) has proven to be a powerful technique for NMT \cite{kim-rush-2016-sequence,wang2021selective} to transfer knowledge from the teacher model to student models.
In particular, we first use the teacher models to generate synthetic corpus in the forward direction (i.e., En$\to$Zh). Then, we use the generated corpus to train our student models. 

Notably, Right-to-Left (R2L) knowledge distillation is a good complement to the Left-to-Right (L2R) way and can further improve model performance.

\subsubsection{Forward-Translation}
Using monolingual data from the source language to further enhance the performance and robustness of the model is also an effective approach. We use the ensemble model to generate high quality forward-translation data and obtain a stable improvement in En$\to$Zh and En$\to$De directions.

\subsection{Iterative In-domain Knowledge Transfer}
\label{sec:ind-trans}

Since in-domain knowledge transfer \cite{meng2020wechat} delivered a massive performance boost last year, we still use this technique in En$\leftrightarrow$Ja and En$\to$De this year. It is not applied to En$\to$Zh because no significant improvement is observed. We guess the reason is that the in-domain finetuning in the En$\to$Zh direction does not bring a significant improvement compared to the other directions. And in-domain knowledge transfer is aiming at enhancing the effect of finetuning, so this does not have a noticeable effect in the English-Chinese direction.

We first use normal finetuning in Sec.~\ref{sec:finetune} to equip our models with in-domain knowledge.
Then, we ensemble these models to translate the source monolingual data into the target language.
We use 4 models with different architectures and training data as our ensemble model.
Next, we combine the source language sentences with the generated in-domain target language sentences as pseudo-parallel corpus.
Afterwards, we retrain our models with both in-domain pseudo-parallel data and general domain synthetic data.

\subsection{Data Augmentation}

Once the pseudo-data is constructed, we further obtain diverse data by adding different noise.
Compared to previous years' WMT competitions, we implement a multi-level static noise approach for our pseudo corpus:
\begin{itemize}
\item Token-level: Noise on every single subword after byte pair encoding.
\item Word-level: Noise on every single word before byte pair encoding.
\item Span-level: Noise on a continuous sequence of tokens before byte pair encoding.
\end{itemize}

The different granularities of noise make the data more diverse. The noise types are random replacement, random deletion and random permutation. We apply the three noise types in a parallel way for each sentence. The probability of enabling each of the three operations is 0.2.

Furthermore, an on-the-fly noise approach is applied to the synthetic data. By using on-the-fly noise, the model is trained with different noises in every epoch rather than all the same along this training stage.

\subsection{In-domain Finetuning}
\label{sec:finetune}

A domain mismatch exists between the obtained system trained with large-scale general domain data and the target test set. To alleviate this mismatch, we finetune these convergent models on small scale in-domain data, which is widely used for domain adaption \cite{luong2015stanford, lietalniutrans2019}. We take the previous test sets as in-domain data and extract documents that are originally created in the source language for each translation direction \cite{baiduwmt2019}. We also explore several advanced finetuning approaches to strengthen the effects of domain adaption and ease the exposure bias issue, which is more serious under domain shift.

\paragraph{Target Denoising \cite{meng2020wechat}.}
In the training stage, the model never sees its own errors. Thus the model trained with teacher-forcing is prune to accumulated errors in testing \cite{ranzato2015exposurebias}. To mitigate this training-generation discrepancy, we add noisy perturbations into decoder inputs when finetuning. Thus the model becomes more robust to prediction errors by target denoising.
Specifically, the finetuning data generator chooses 30\% of sentence pairs to add noise, and keeps the remaining 70\% of sentence pairs unchanged. For a chosen pair, we keep the source sentence unchanged, and replace the $i$-th token of the target sentence with (1) a random token of the current target sentence 15\% of the time (2) the unchanged $i$-th token 85\% of the time.

\paragraph{Graduated Label-smoothing \cite{wang2020inference}.}
Finetuning on a small scale in-domain data can easily lead to the over-fitting phenomenon which is harmful to the model ensemble. It generally appears as the model over confidently outputting similar words. To further preventing over-fitting of in-domain finetuning, we apply the Graduated Label-smoothing approach, which assigns a higher smoothing penalty for high-confidence predictions, during in-domain finetuning.
Concretely, following the paper's setting, we set the smoothing penalty to 0.3 for tokens with confidence above 0.7, zero for tokens with confidence below 0.3, and 0.1 for the remaining tokens.

\paragraph{Confidence-Aware Scheduled Sampling.} Vanilla scheduled sampling~\cite{zhang-etal-2019-bridging} simulates the inference scene by randomly replacing golden target input tokens with predicted ones during training. However, its critical schedule strategies are only based on training steps, ignoring the real-time model competence. To address this issue, we propose confidence-aware scheduled sampling \cite{liu2021confidence}, which quantifies real-time model competence by the confidence of model predictions. At the t-th target token position, we calculate the model confidence $conf(t)$ as follow:
\begin{equation}
    conf(t) = P(y_t|{\rm \bf y_{<t}, \bf X, \theta})
\end{equation}
Next, we design fine-grained schedule strategies based on the model competence. The fine-grained schedule strategy is conducted at all decoding steps simultaneously:
\begin{equation}
    y_{t-1} = \begin{cases}
    y_{t-1}\qquad \ if\ conf(t) \leq t_{golden} \\
    \hat{y}_{t-1}\qquad \ else \\
    \end{cases}
\end{equation}
where $t_{golden}$ is a threshold to measure whether $conf(t)$ is high enough ($e.g.,$ 0.9) to sample the predicted token $\hat{y}_{t-1}$.

We further sample more noisy tokens at high-confidence token positions, which prevents scheduled sampling from degenerating into the teacher forcing mode.
\begin{equation}
    y_{t-1} = \begin{cases}
    y_{t-1}\quad \ if\ conf(t) \leq t_{golden} \\
    \hat{y}_{t-1}\quad \ if\ t_{golden} < conf(t) \leq t_{rand} \\
    y_{rand}\quad if\ conf(t) > t_{rand}
    \end{cases}
\end{equation}
where $t_{rand}$ is a threshold to measure whether $conf(t)$ is high enough ($e.g.,$ 0.95) to sample the random target token $\hat{y}_{rand}$.

\paragraph{Scheduled Sampling Based on Decoding Steps.} We propose scheduled sampling methods based on decoding steps from the perspective of simulating the distribution of real translation errors \cite{liuemnlp2021scheduled}. Namely, we gradually increase the selection probability of predicted tokens with the growth of the index of  decoded tokens. At the $t$-th decoding step, the probability of sampling golden tokens $g(t)$ is calculated as follow:

\begin{itemize}
\item Linear Decay: $g(t) = \max(\epsilon , kt + b)$, where $\epsilon$ is the minimum value, and $k < 0$ and $b$ is respectively the slope and offset of the decay.
\item Exponential Decay: $g(t) = k^t$, where $k < 1$ is the radix to adjust the decay.
\item Inverse Sigmoid Decay: $g(t) = \frac{k}{k + e^{\frac{t}{k}}}$, where $e$ is the mathematical constant, and $k \geq 1$ is a hyperparameter to adjust the decay. 
\end{itemize}
Following our preliminary conclusions \cite{liuemnlp2021scheduled}, we choose the exponential decay and set $k$ to 0.99 by default.

\begin{algorithm}[t!]
	\renewcommand{\algorithmicrequire}{\textbf{Input:}}
	\renewcommand{\algorithmicensure}{\textbf{Output:}}
	\caption{Boosted Self-BLEU based Ensemble}
	\label{alg:1}
	\begin{algorithmic}[0]
        \REQUIRE 
        \STATE List of candidate models M = \{$m_0$, ..., $m_n$\} \\
        \STATE Valid set BLEU for each model B = \{$b_i$, ..., $b_n$\} \\
        \STATE Average Self-BLEU for each model S = \{$s_i$, ..., $s_n$\} \\
        \STATE The number of models $n$ \\
        \STATE The number of ensemble models $c$ \\
		\ENSURE Model combinations C
	\end{algorithmic}
	\begin{algorithmic}[1]
		\FOR{$i \leftarrow 1 $ $to$ $n$}
        \STATE $score_i$ = 
        \begin{small}
        $(b_i - min(B))\cdot weight+(max(S) - s_i)$
        \end{small}
        \STATE $weight$ = $\frac{(max(S)-min(S))}{(max(B)-min(B))}$
        \ENDFOR
		\STATE Add the highest score model to candidates list C = \{  $m_{top}$ \} \\
        \WHILE {$|C|$ \textless \;$c$ }
        \STATE $ index$ = 
        \begin{small}
        $\mathop{\arg\min}\limits_{i} \frac{1}{|M-C|}$
        $\sum\limits_{i \in {M-C},j\in{C}} BLEU(i,j)$
        \end{small}
        \STATE Add $m_{index}$ to candidate list C
        \ENDWHILE
		\STATE \textbf{return} C
    \end{algorithmic}
\end{algorithm}

\subsection{Boosted Self-BLEU based Ensemble (BSBE)}
\label{sec:ens}
After we get numerous finetuned models, we need to search for the best combination for ensemble model. Ordinary random or greedy search is oversimplified to search for a good model combination and enumerate over all combinations of candidate models is inefficient. The Self-BLEU based pruning strategy \cite{meng2020wechat} we proposed in last year's competition achieve definite improvements over the ordinary ensemble. 

However, diversity is not the only feature we need to consider but the performance in the valid set is also an important metric. Therefore, we combine Self-BLEU and valid set BLEU together to derive a Boosted Self-BLEU-based Ensemble (BSBE) algorithm. Then, we apply a greedy search strategy in the top N ranked models to find the best ensemble models. 

See algorithm \ref{alg:1} for the pseudo-code. The algorithm takes as input a list of n strong single models M, BLEU scores on valid set for each model B, average Self-BLEU scores for each model S, the number of models $n$ and the number of ensemble models $c$. The algorithm return a list C consists of selected models. We calculate the weighted score for each model as line 2 in the pseudo-code. The weight calculated in line 3 is a factor to balance the scale of Self-BLUE and valid set BLEU. Then the list C initially contains the model $m_{top}$ has a highest weighted score. Next, we iteratively re-compute the average Self-BLEU between the remaining models in $|M-C|$ and selected models in C, based on which we select the model has minimum Self-BLEU score into C.

In our experiments, we save around 95\% searching time by using this novel method to achieve the same BLEU score of the Brute Force search. We will further analyze the effect of Boosted Self-BLEU based Ensemble in section \ref{self_bleu_ens}.

\section{Experiments And Results}  \label{experiments}
\subsection{Settings}
The implementation of our models is based on Fairseq\footnote{https://github.com/pytorch/fairseq} for En$\to$Zh and EN$\to$De, and OpenNMT\footnote{https://github.com/OpenNMT/OpenNMT-py} for En$\leftrightarrow$Ja.
All the single models are carried out on 8 NVIDIA V100 GPUs, each of which has 32 GB memory. 
We use the Adam optimizer with $\beta_{1}$ = 0.9, $\beta_{2}$ = 0.998. 
The gradient accumulation is used due to the high GPU memory consumption.
The batch size is set to 8192 tokens per GPU and we set the “update-freq” parameter in Fairseq to 2.
The learning rate is set to 0.0005 for Fairseq and 2.0 for OpenNMT. We use warmup step = 4000.
We calculate sacreBLEU\footnote{https://github.com/mjpost/sacrebleu} score for all experiments which is officially recommended.

\begin{table}[!t]
\centering
\scalebox{0.90} {
\begin{tabular}{l|r|c|c}
\hline
 & En$\to$Zh & En$\to$De & En$\leftrightarrow$Ja  \\
\hline
Bilingual  Data & 30.7M  & 74.8M & 12.3M \\
Source Mono Data & 200.5M & 332.8M & 210.8M \\
Target Mono Data & 405.2M & 237.9M & 354.7M \\
\hline
\end{tabular}
}
\caption{Statistics of all training data.} 
\label{t:train_data}
\end{table}

\begin{table*}[t!]
\centering
\scalebox{0.92} {
\begin{tabular}{l| c | c | c | c }
\hline
\bf \textsc{System} & En$\to$Zh & En$\to$Ja & Ja$\to$En & En$\to$De \\
\hline
Baseline  & 44.53 & 35.78 & 19.71 & 33.28 \\
+ Back Translation & 46.52 & 36.12 & 20.82 & 35.28 \\
+ Knowledge Distillation & 47.14 & 36.66 & 21.63 & 36.38 \\
+ Forward Translation & 47.38 & -- & -- & 36.78 \\
+ Mix BT & 48.17 & 37.22 & 22.11 & -- \\
~~~~ + \emph{Finetune} & 49.81  & 42.54 & 25.91 & 39.21 \\
~~~~ + \emph{Advanced Finetune} & 50.20 & -- & -- & 39.56 \\
\hline
+ 1st In-domain Knowledge Transfer & -- & 40.32 & 24.49 & 39.23\\
~~~~ + \emph{Finetune} & -- & 43.66 & 26.24 & -- \\
~~~~ + \emph{Advanced Finetune} & -- & -- & -- & 39.87\\
+ 2nd In-domain Knowledge Transfer & -- & 43.69 & 25.89 & --\\
~~~~ + \emph{Finetune} & -- & 44.23 & 26.27 & --\\
~~~~ + \emph{Advanced Finetune} & --& 44.42 & 26.38 & --\\
\hline
+ Normal Ensemble & 50.57 & 45.11 & 28.01 & 40.42 \\
\hline
+ BSBE & {\ \ \ \bf 50.94 $\star$} & \ \ \ \bf 45.35 $\star$ & \ \ \ \bf 28.24 $\star$ & 40.59\\
\hline
+ Post-Process & -- & -- & -- & \ \ \ \bf 41.88 $\star$ \\
\hline
\end{tabular}
}
\caption{Case-sensitive BLEU scores (\%) on the four directions \emph{newstest2020}, where `$\star$' denotes the submitted system. Mix BT means we use multiple parts of Back Translation data with different generation strategies. The \emph{Advanced Finetune} methods outperform the normal \emph{Finetune} and we report the best results in single model. BSBE outperforms Normal Ensemble in all four directions.
} 
\label{t:encn}
\end{table*}

\subsection{Dataset}
The statistics of all training data is shown in Table \ref{t:train_data}. For each language pair, the bilingual data is the combination of all parallel data released by WMT21. For monolingual data, we select data from News Crawl, Common Crawl and Extended Common Crawl, it is then divided into several parts, each containing 50M sentences. 

For general domain synthetic data, we use all the target monolingual data to generate back-translation data and a part of source monolingual data (about 80 to 100 million for different languages) to get forward translation data. For the in-domain pseudo-parallel data, we use the entire source monolingual data and bilingual data. All the test and valid data from previous years are used as in-domain data.

We use the methods described in Sec.~\ref{sec:data_filter} to filter bilingual and monolingual data.

\subsection{Pre-processing and Post-processing}
English and German sentences are segmented by Moses\footnote{http://www.statmt.org/moses/}, while Japanese use Mecab\footnote{https://github.com/taku910/mecab} for segmentation. We segment the Chinese sentences with an in-house word segmentation tool. We apply punctuation normalization in English, German and Chinese data. Truecasing is applied to English$\leftrightarrow$ Japanese and English$\to$German.
We use byte pair encoding BPE \cite{sennrichetal2016bpe} with 32K operations for all the languages.

For the post-processing, we apply de-truecaseing and de-tokenizing on the English and German translations with the scripts provided in Moses. For the Chinese translations, we transpose the punctuations to the Chinese format.

\subsection{English$\to$Chinese}
The results of En$\to$Zh on newstest2020 are shown in Table \ref{t:encn}.
For the En$\to$Zh task, filtering out part of sentence pairs containing English characters in Chinese sentences shows a significant improvement in the valid set. After applying large-scale Back-Translation, we obtain +2.0 BLEU score on the baseline.
We further gain +0.62 BLEU score after applying knowledge distillation and +0.24 BLEU from Forward-Translation. Surprisingly, we observe that adding more BT data from different shards with different generation strategy can further boost the model performance to 48.17.
The finetuned model achieves a 49.81 BLEU score, which demonstrates that the domain of the training corpus is apart from the test set domain. 
The \textit{advanced} finetuning further brings about 0.41 BLEU score gains compared to normal finetune. Our best single model achieves a 50.22 BLEU score.

In preliminary experiments, we select the best performing models as our ensemble combinations obtaining +0.4 BLEU score. On top of that, even after searching hundreds of models, no better results are obtained. With BSBE strategies in Sec.~\ref{sec:ens}, a better model combination with less number of models are quickly searched, and we finally achieve 50.94 BLEU score. Our WMT2021 English$\to$Chinese submission achieves a SacreBLEU score of 36.9, which is the highest among all submissions and chrF score of 0.337.

\subsection{English$\to$Japanese}
The results of En$\to$Ja on newstest2020 are shown in Table \ref{t:encn}.
For the En$\to$Ja task, we filter out the sentence pairs containing Japanese characters in the English side and vice versa. The Back-Translation and Knowledge Distillation improve the baseline from 35.78 to 36.66. Adding more BT data further brings in 0.56 improvements. The improvement by finetuning is much larger than other directions, which is 5.32 BLEU. 
We speculate that this is because there is less bilingual data for English and Japanese than for other languages, and the test results for Japanese are char level BLEU so this direction is more influenced by the in-domain finetuning.
Two In-domain knowledge transfers improve BLEU score from 37.22 to 43.69. Normal finetune  still provides 0.54 improvements after in-domain knowledge transfer. Then, we apply \textit{advanced} finetuning methods to further get 0.19 BLEU improvements. Our final ensemble result outperforms baseline 9.57 BLEU.

\begin{table*}[t!]
\centering
\scalebox{0.92} {
\begin{tabular}{l| c| c | c | c}
\hline
\bf \textsc{Model} & \bf \textsc{En-Zh} & \bf \textsc{En-Ja}  & \bf \textsc{Ja-En} & \bf \textsc{En-De} \\
\hline
Transformer                     & 49.92 & 44.27 & 26.12 & 39.76 \\
Transformer with Post-Norm      & 49.97 & -     & -     & - \\
Average Attention Transformer   & 49.91 & 44.38  & 26.31 & 39.62 \\
Weighted Attention Transformer  & 49.99 & -     & -     & 39.74 \\
Average First Transformer $\ast$  & 50.14 & \bf 44.42 & 26.37 & \bf 39.87 \\
Average Bottom Transformer $\ast$ & 50.10 & 44.36 & \bf 26.38 & 39.77 \\
Dual Attention Transformer $\ast$ & \bf 50.20 & -     & -     & \bf 39.87 \\
Talking-Heads Attention         & 49.89 & -     & -     & 39.70 \\
\hline
\end{tabular}
}
\caption{Case-sensitive BLEU scores (\%) on the four translation directions \emph{newstest2020} for different architecture. The model with `$\ast$' is the Mixed-AAN variants. The \textbf{bolded} scores correspond to the best single model scores in Table \ref{t:encn}.
} 
\label{t:model_results}
\end{table*}

\begin{small}
\begin{table*}[t!]
\centering
\scalebox{0.92} {
\begin{tabular}{l| c| c | c | c | c | c | c | c }
\hline
\bf \textsc{Model} & Transformer & Post-Norm  & AAN & Weighted & Avg-First & Self-First & Dual & TH \\
\hline
Transformer     & 100   & 78.12 & 76.02 & 75.08 & 74.47 & 74.02 & 73.51 & 72.63 \\
Post-Norm       & 78.12 & 100   & 76.12 & 75.10 & 74.33 & 74.05 & 73.45 & 72.59 \\
AAN             & 76.02 & 76.12 & 100   & 79.24 & 74.81 & 74.97 & 73.43 & 72.13 \\
Weighted        & 75.08 & 75.10 & 79.24 & 100   & 74.72 & 74.93 & 73.55 & 72.21 \\
Avg-First $\ast$ & 74.46 & 74.33 & 74.81 & 74.72 & 100   & 75.25 & 74.28 & 72.25 \\
Avg-Bot $\ast$  & 74.02 & 74.05 & 74.97 & 74.93 & 75.25 & 100   & 74.21 & 72.33 \\
Dual $\ast$     & 73.51 & 73.45 & 73.43 & 73.55 & 74.28 & 74.21 & 100   & 72.23 \\
TH              & \bf 72.63 & \bf 72.59 & \bf 72.13 & \bf 72.21 & \bf 72.25 & \bf 72.33 & \bf 72.23 & 100 \\
\hline
\end{tabular}
}
\caption{Self-BLEU scores (\%) between different architectures. For simplicity, we refer to these models as Transformer (Pre-Norm Transformer), Post-Norm (Post-Norm Transformer), AAN (Average Attention Transformer), Weighted (Weighted Attention Transformer), Avg-First (Average First Transfromer), Avg-Bot (Average Bottom Transformer), Dual (Dual Attention Transformer), TH (Talking-Heads Attention). The model with `$\ast$' is the MixAAN variants.
} 
\label{t:model_self_bleu}
\end{table*}
\end{small}

\subsection{Japanese$\to$English}
The Ja$\to$En task follows the same training procedure as En$\to$Ja.
From Table \ref{t:encn}, we can observe that Back-Translation can provide 1.11 BLEU improvements from baseline. Knowledge Distillation and more BT data can improve the BLEU score from 20.82 to 22.11. The finetuning improvement is 3.8 which is slightly less than the En$\to$Ja direction but still larger than En$\to$Zh and En$\to$De. We also apply two-turn in-domain knowledge transfer and further boost the BLEU score to 25.89. After normal finetuning, the BLEU score achieves 26.27. The \textit{advanced} finetuning methods provide a slight improvement on Ja$\to$En. After ensemble, we achieve 28.24 BLEU in newstest2020.

\subsection{English$\to$German}
The results of En$\to$De on newstest2020 are shown in Table \ref{t:encn}. 
After adding back-translation, we improve the BLEU score from 33.28 to 35.28. Knowledge Distillation further boosts the BLEU score to 36.58. The finetuning further brings in 2.63 improvements.
After injecting the in-domain knowledge into the monolingual corpus, we get another 0.31 BLEU gain. We apply a post-processing procedure on En$\to$De. Specifically, we normalize the English quotations to German ones in German hypotheses, which brings in 1.3 BLEU improvements.

\section{Analysis}
\label{analysis}
To verify the effectiveness of our approach, we conduct analytical experiments on model variants, finetune methods, and ensemble strategies in this section.

\subsection{Effects of Model Architecture} \label{effect_of_model}
We conduct several experiments to validate the effectiveness of Transformer \cite{transformer2017} variants we used and list results in Table \ref{t:model_results}. We also investigate the diversity of different variants and the impacts on the model ensemble. The results is listed in Table \ref{t:model_self_bleu} and Table \ref{t:model_ensemble}. Here we take En$\to$Zh models as examples to conduct the diversity and ensemble experiments. The results in other directions show similar trends.

\paragraph{Performance.}
As shown in Table \ref{t:model_results}, AAN performs slightly worse than other variants in En$\to$Zh but Mixed-AAN variants outperform normal Transformer. Weighted Attention Transformer provides noticeable improvement compare to AAN and sometimes better than vanilla Transformer.
\paragraph{Diversity.}
The Self-BLEU scores in Table \ref{t:model_self_bleu} demonstrate the difference between two models, more different models generally have lower scores. As we can see, AAN and all the variants with AAN have an absolutely lower Self-BLEU score with the Transformer. The Talking-Heads Attention has the minimum scores among all the variants.
\paragraph{Ensemble.}
In our preliminary experiments, we observe that more diverse models can significantly help the model ensemble. The results are listed in Table \ref{t:model_ensemble}. We get a more robust ensemble model with only four models using our novel variants than searching from dozens of Deeper and Wider Transformer models. Even these four models are trained with the same training data. After we combine the four models with Deeper and Wider Transformer, we can further get a significant improvement.

Take En$\to$Zh as an examble, our final submission consist of 1 Average First Transformer, 1 Average Bottom Transformer, 1 Dual Attention Transformer, 1 Weighted Attention Transformer and 1 Transformer with Post-Norm.

\begin{small}
\begin{table}[t!]
\centering
\scalebox{0.92} {
\begin{tabular}{l| c }
\hline
\bf \textsc{Models} & newstest2020 \\
\hline
Deeper \& Wider Transformer    & 50.31  \\
Weighted \& Mixed-AAN        & 50.44 \\
Ensemble with all models above     & \bf 50.62 \\
\hline
\end{tabular}
}
\caption{Ensemble results with different architectures. The first row is the ensemble results with 10 deeper and wider models searched from dozens of ones. The second row is the ensemble results with only 4 Weighted Attention Transformer and Mixed-AAN models. 
} 
\label{t:model_ensemble}
\end{table}
\end{small}

\subsection{Effects of Boosted Self-BLEU based Ensemble} \label{self_bleu_ens}
To verify the superiority of our Boosted Self-BLEU based Ensemble (BSBE) method, we randomly select 10 models with different architecture and training data. For our submitted system, we search from over 500 models. We use a greedy search algorithm \cite{deng2018alibaba} as our baseline. The greedy search greedily selects the best performance model into candidate ensemble models. If the selected model provides a positive improvement, we keep it in the candidates. Otherwise, it is added to a temporary model list and still has a weak chance to be reused in the future. One model from the temporary list can be reused once, after which it is withdrawn definitely. We compare the results of greedy search, BSBE and Brute Force and list the ensemble model BLEU and the number of searches in Table \ref{t:self_bleu_ensemble}. Note that $n$ is the number of models, which is 10 here. For BSBE, we need to get the translation result of every model to calculate the Self-BLEU. After that, we only need to perform the inference process once.

\subsection{Effects of Advanced Finetuning}
In this section, we describe our experiments on advanced finetuning in the four translation directions. As shown in Table \ref{t:finetune_results}, all the advanced finetuning methods outperform normal finetuning. 
For En$\to$Zh, Scheldule Sampling Based on Decoding Steps with Graduated Label Smoothing improves the model performance from 49.81 to 50.20. For En$\leftrightarrow$Ja, Target Denoising with Graduated Label Smoothing provides the highest BLEU gain, which are 0.19 and 0.11. For the En$\to$De direction, Confidence-Aware Scheldule Sampling with Graduated Label Smoothing performs the best, improving from 39.21 to 39.42.
These findings are in line with the conclusion of \citet{wangsennrichexposure2020} that links exposure bias with domain shift. 

\begin{tiny}
\begin{table}[t!]
\centering
\scalebox{0.92} {
\begin{tabular}{l| c | c}
\hline
\bf \textsc{Algorithm} & BLEU & Number of Searches \\
\hline
Greedy    & 50.19 & $2n$ \\
Brute Force      & 50.44 & $\sum_{i=1}^n C_{n}^{i}$ \\
BSBE    & \bf 50.44 & $n+1$ \\
\hline
\end{tabular}
}
\caption{Results of different search algorithm. $n$ is the total number of models used for the search. The number of searches is number that the methods need to translate the valid set. Our BSBE achieves comparable BLEU score as Brute Force search and significantly reduces the searching time.
} 
\label{t:self_bleu_ensemble}
\end{table}
\end{tiny}

\begin{table*}[t!]
\centering
\scalebox{0.92} {
\begin{tabular}{l| c| c | c | c}
\hline
\bf \textsc{Finetuning Approach} & \bf \textsc{En-Zh} & \bf \textsc{En-Ja}  & \bf \textsc{Ja-En} & \bf \textsc{En-De} \\
\hline
Normal  & 49.81  & 44.23 &  26.27  & 39.21 \\
Graduated Label Smoothing & 49.95 & 44.32 & 26.35 & 39.32 \\
~~~~ + \emph{Target Denoising}  & 50.09 & \bf 44.42 & \bf 26.38 & 39.34 \\
~~~~ + \emph{Confidence-Aware Scheldule Sampling} & 50.17 & 44.35 & 26.33 & \bf 39.42 \\
~~~~ + \emph{Scheldule Sampling Based on Decoding Steps} & \bf 50.20 & 44.36 & 26.33 & 39.40 \\
\hline
\end{tabular}
}
\caption{Case-sensitive BLEU scores (\%) on the four translation directions \emph{newstest2020} for different finetuning approaches. 
We report the highest score and bold the best result among different finetuning approaches.
} 
\label{t:finetune_results}
\end{table*}

\section{Conclusion}  \label{conclusion}
We investigate various novel Transformer based architectures to build robust systems. Our systems are also built on several popular data augmentation methods such as back-translation, knowledge distillation and iterative in-domain knowledge transfer. We enhance our system with advanced finetuning approaches, i.e., target denoising, graduated label smoothing and confidence-aware scheduled sampling. A boosted Self-BLEU based model ensemble is also employed which plays a key role in our systems.
Our constrained systems achieve 36.9, 46.9, 27.8 and 31.3 case-sensitive BLEU scores on English$\to$Chinese, English$\to$Japanese, Japanese$\to$English and English$\to$German, respectively. The BLEU scores of English$\to$Chinese, English$\to$Japanese and Japanese$\to$English are the highest among all submissions, and that of English$\to$German is the highest among all constrained submissions. 

\section*{Acknowledgements}
Yijin Liu and Jinan Xu have been supported by the National Key R\&D Program of China (2020AAA0108001) and the National Nature Science Foundation of China (No. 61976015, 61976016, 61876198 and 61370130). The authors would like to thank the anonymous reviewers for their valuable comments and suggestions to improve this paper. 

\bibliography{anthology_rebiber,custom}
\bibliographystyle{acl_natbib}

\end{document}